\documentclass[10pt,twocolumn,letterpaper]{article}

\usepackage[pagenumbers]{wacv} 

\usepackage[pagebackref,breaklinks,colorlinks]{hyperref}
\usepackage{url}

\usepackage[utf8]{inputenc} 
\usepackage[T1]{fontenc}    
\usepackage{hyperref}       
\usepackage{graphicx}
\usepackage{url}            
\usepackage{booktabs}       
\usepackage{amsfonts}       
\usepackage{nicefrac}       
\usepackage{microtype}      
\usepackage{xcolor}         

\usepackage{multicol}
\usepackage{multirow}
\usepackage{pifont}
\usepackage{caption}
\usepackage{amsmath}
\usepackage{listings}
\usepackage[inline]{enumitem}
\usepackage{algorithm}
\usepackage{color, colortbl}

\usepackage{url}
\usepackage{booktabs}       
\usepackage{nicefrac}       
\usepackage{microtype}      
\usepackage{xcolor}         
\usepackage{epsfig}
\usepackage{times}
\usepackage{rotating}
\usepackage{multirow}
\usepackage{tabulary}
\usepackage{float}
\usepackage{textcomp}

\usepackage{caption}
\usepackage{verbatim}
\usepackage{wrapfig}

\usepackage{hyperref}
 \hypersetup{
     allcolors = magenta
     }
     
\def \pzo {\phantom{}} 
\def \dzo {\phantom{}}

\definecolor{lime}{RGB}{205, 237, 250}

\definecolor{pink}{RGB}{245,220,250}
\definecolor{lemon}{RGB}{255, 255, 171}

\newcommand{\ignore}[1]{}

\usepackage[pagebackref,breaklinks,colorlinks]{hyperref}

\usepackage[capitalize]{cleveref}
\crefname{section}{Sec.}{Secs.}
\Crefname{section}{Section}{Sections}
\Crefname{table}{Table}{Tables}
\crefname{table}{Tab.}{Tabs.}

\def\wacvPaperID{1929} 
\def\confName{WACV}
\def\confYear{2024}

\begin{document}

\title{SimA: Simple Softmax-free Attention for Vision Transformers }

\author{
 \fontsize{11}{11} \selectfont Soroush Abbasi Koohpayegani $\quad$ Hamed Pirsiavash   \vspace{1.5mm}\\
\fontsize{11}{11} \selectfont University of California, Davis\\
{\fontsize{9}{9} \selectfont \texttt{soroush@ucdavis.edu \quad hpirsiav@ucdavis.edu}}
}

\maketitle

\begin{abstract}
   Recently, vision transformers have become very popular. However, deploying them in many applications is computationally expensive partly due to the Softmax layer in the attention block. We introduce a simple yet effective, Softmax-free attention block, SimA, which normalizes query and key matrices with simple $\ell_1$-norm instead of using Softmax layer. Then, the attention block in SimA is a simple multiplication of three matrices, so SimA can dynamically change the ordering of the computation at the test time to achieve linear computation on the number of tokens or the number of channels. We empirically show that SimA applied to three SOTA variations of transformers, DeiT, XCiT, and CvT, results in on-par accuracy compared to the SOTA models, without any need for Softmax layer. Interestingly, changing SimA from multi-head to single-head has only a small effect on the accuracy, which further simplifies the attention block. Moreover, we show that SimA is much faster on small edge devices, e.g., Raspberry Pi, which we believe is due to higher complexity of Softmax layer on those devices. The code is available here: \textcolor{magenta}{\href{https://github.com/UCDvision/sima}{https://github.com/UCDvision/sima}}
\end{abstract}

\section{Introduction}
Recently, vision transformers have become very popular. Compared to CNNs, they achieve better accuracy, however, deploying transformers in devices with smaller computational resources is challenging. One reason is that a transformer model calls the Softmax layer several times which calls $exp(.)$ operation consequently. We know that the $exp(.)$ operation is costly particularly in smaller devices with limited computational resources. For instance, implementing $exp(.)$ on FGPA is much more costly compared to implementing simple multiplication or addition operations.

As an example observation, Table A1 of \cite{ivanov2021data} measures the run-time of each component for a BERT encoder on V100 GPUs. Softmax consumes more time compared to any other components including query ($Q$), key ($K$), value ($V$) operation (Softmax: 453 $\mu s$ , $QKV$ projections: 333 $\mu s$, $QK^T$: 189 $\mu s$). This is remarkable since the FLOPS of Softmax is much lower than those other components (Softmax: 0.2 GFLOPS, $QKV$ projections: 25.7 GFLOPS, $QK^T$: 4.3 GFPLOS). Similar observation are made in \cite{stevens2021softermax,vasyltsov2021efficient}.

We are interested in simplifying the attention mechanism by removing the Softmax layer. We believe one role of the Softmax layer is to normalize the attention values so that tokens can compete with each other. Our main idea is to enable this competition by normalizing the query and key matrices with their $\ell_1$-norm before multiplying them. Then, removing the Softmax layer results in the whole attention mechanism to boil down to simply multiplying three matrices ``query'', ``key'', and ``value''. While $\ell_1$-norm has been used in transformers before \cite{guo2022beyond}, the way we are using it to simplify the computational flow of the transformer is novel. 

As a bi-product, due to the associative property of multiplication, there are two possible orderings of multiplying these three matrices at the test time. Depending on the ordering, the computation can be quadratic on the number of tokens, $N$, or that of channels, $D$. Hence, we can reduce the computation further by dynamically deciding on the ordering at the test time by comparing $N$ and $D$ without affecting the training process. Moreover, since we normalize the vectors before multiplying, our method is numerically more stable so we use half-precision floating point without overflowing. 

\begin{figure*}[t]

    \centering
    \begin{minipage}{0.45\textwidth}
        \centering
        \includegraphics[width=1.0\linewidth]{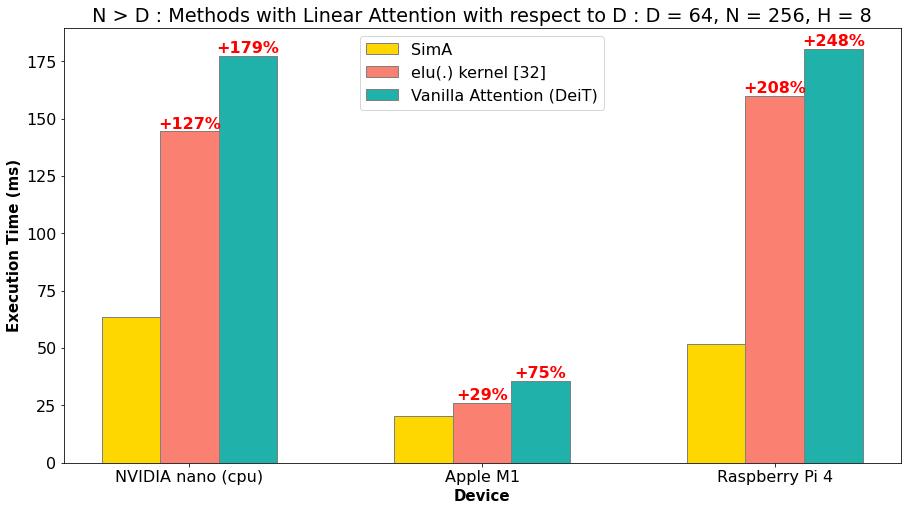}
    \end{minipage}%
    \begin{minipage}{0.45\textwidth}
        \centering
        \includegraphics[width=1.0\linewidth]{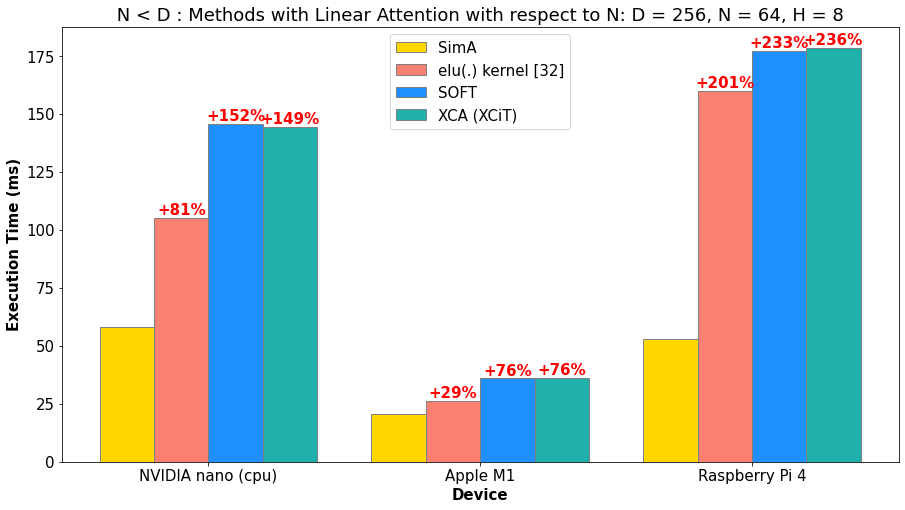}
    \end{minipage}
    \caption{\textbf{Comparison on Edge devices:} We evaluate performance of a single attention block for each model on 3 different devices: Raspberry Pi 4 (Quad core Cortex-A72 @ 1.5GHz), NVIDIA Jetson Nano (Quad-core ARM A57 @ 1.43 GHz), and Apple M1. To measure the effect of $exp(.)$ only, we fix the order of ($QK^TV$) product so that all models have the same dot product complexity. We set $N>D$ for left and $N<D$ for the right plots. We repeat average of the execution time over $1000$ runs. We observe that SimA is faster than other methods, which we believe is due to the increased complexity of $exp(.)$ operation compared to $\ell_1$ normalization on edge devices.}
    \label{fig:softmax_cost}
    \vspace{-.1in}
\end{figure*}


The attention mechanism deals with the tokens without considering their ordering. This is an interesting property that opens the door to many applications. For instance, the distribution of the tokens is relatively robust compared to CNNs when we mask (drop) 75\% of the tokens in masking auto-encoder (MAE \cite{he2021masked}). Moreover, the tokens can be seen as a non-ordered set that can come from various sources (e.g., multiple cameras or non-camera sensors). Note that this permutation equivariance property does not exist in some other models like MLP-Mixer \cite{tolstikhin2021mlp}. Hence, instead of using MLP-Mixer that does not have Softmax by default, we are interested in removing Softmax from the original transformers to keep this permutation equivariance property.

We perform experiments with our simple attention block, denoted SimA, by using it in standard vision transformers, DeiT, CvT, and XCiT. SimA achieves on-par results with SOTA on ImageNet classification, MS-COCO object detection and segmentation, and also self-supervised learning.

In summary, our SimA attention block does not use Softmax, which makes it computationally efficient generally (see Fig.~\ref{fig:softmax_cost} and Table \ref{tab:exec_time}), and on the edge devices specifically. SimA can dynamically choose to be linear on $N$ or $D$ at the test time depending on the image resolution or the number of tokens. Changing Multi-head attention to Single-head one or changing GELU activation function to ReLU, has a very small effect on the accuracy of SimA. This makes SimA simple and effective for various applications.

\section{Method}

\subsection{Background on Vision Transformers:} 
\textbf{Self-Attention Block:} The original vision  transformer \cite{dosovitskiy2020image} uses the self-attention block introduced in  \cite{vaswani2017attention}. Self-attention block gets $X \in \mathbb{R}^{N\times D}$ as the input where $N$ is the number of tokens and $D$ is the dimensionality of each token. Then $W_q \in \mathbb{R}^{D\times D}$, $W_k\in \mathbb{R}^{D\times D}$ and $W_v\in \mathbb{R}^{D\times D}$ projects $X$ into three $N\times D$ matrices: query ($Q = XW_q$), key ($K = XW_k$) and value ($V = XW_v$). We calculate attention matrix $A \in \mathbb{R}^{N\times N}$ defined as $A=Softmax(QK^T/\sqrt{D})$ where $Softmax$ is applied to each row independently, so each row in $A$ sums to one. Then, we calculate the output $O = AV$. Each row of $O \in \mathbb{R}^{N\times D}$ corresponds to one token and since rows of $A$ sum to one, each token in a weighted average of the values of all tokens. 

Additionally, Multi-Head Self-Attention (MSA) transformers divide $Q$, $K$, and $V$ of each token into $H$ heads, where each head has its own attention over the corresponding head in all tokens. For example, $Q = [Q_1;Q_2;...;Q_H]$ where $Q_i \in \mathbb{R}^{N\times \frac{D}{H}}$ is the query matrix for the $i$'th head. Then, we calculate $H$ self-attention for all heads in parallel and concatenate the outputs to get $O = [O_1;O_2;...;O_H]$. 
Finally, the self-attention block has an additional output projection $W_{proj}\in \mathbb{R}^{D\times D}$, thus the final output of the self-attention block is $OW_{proj}$ which is of size $\mathbb{R}^{N\times D}$.

\textbf{Cross-covariance Attention Block (XCA):} Vanilla self-attention block has a complexity of $O(DN^2)$ which is quadratic on $N$. \cite{ali2021xcit,shen2021efficient} introduce an attention mechanism that is linear on $N$. In XCA, we calculate the attention matrix with $A=K^TQ$ where $A$ is a $D\times D$ matrix. Next, we apply Softmax on each columns, so that columns sum to one. Then we calculate output as $O=VA$. Note that $A$ is an attention of channels on each other rather than tokens. Compared to vanilla self-attention (MSA), XCA has complexity of $O(D^2N)$. Since XCA is linear on $N$, it is more efficient when $N \gg D$ and it is less efficient when $N \ll D$.  

\textbf{Vision Transformer Block:} Vision transformers architecture contains $n$ consecutive Vision Transformer blocks. Each block has MSA block followed by a Feed-Forward Network (FFN) both with skip connection. FFN is a simple 2-layer MLP which projects tokens from $D$ dimension to $4D$ and again back to $D$ dimensions. FFN uses GELU \cite{hendrycks2016gaussian} as the activation function. Moreover, we use LayerNorm \cite{ba2016layer} on each token before forwarding them through MSA or FFN blocks. The following two updating rules summarize each block of the vision transformer:
\begin{align*}\label{eqn:eq1}
(Step 1)\quad &X \leftarrow X + MSA(\text{LayerNorm}_1(X)) \\
(Step 2)\quad &X \leftarrow X + FFN(\text{LayerNorm}_2(X))
\nonumber
\end{align*}

\begin{figure*}[t]
\centering
\includegraphics[width=0.9\linewidth]{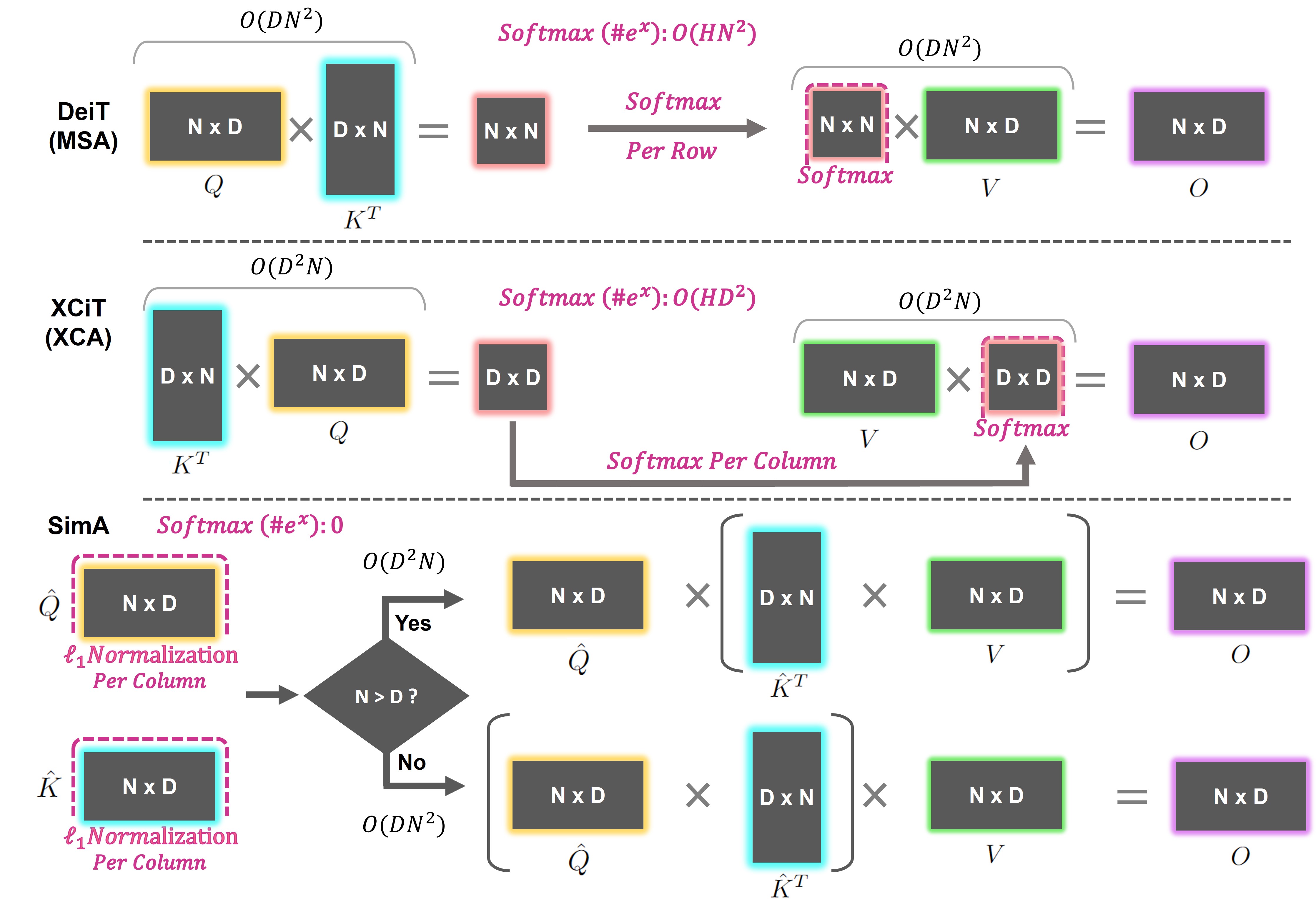}
\vspace{-.15in}

\caption{{\bf Our Simple Attention (SimA):} First, we normalize each channel in $Q$ and $K$ with $\ell_1$-norm across the tokens, 
to get $\hat{Q}$ and $\hat{K}$. Next, we can choose either  $(\hat{Q}\hat{K}^T)V$ or $\hat{Q}(\hat{K}^TV)$ depending on the number of input tokens $N$. Compared to XCA and MSA, our method has following benefits: (1) It is free of Softmax, hence it is more efficient. (2) At test time we can dynamically switch between $(\hat{Q}\hat{K}^T)V$ and $\hat{Q}(\hat{K}^TV)$  based on the number of input tokens (e.g., different image resolution). }
\label{fig:sima}
\vspace{-.1in}
\end{figure*}

\subsection{Simple Attention (SimA):} 
\label{sima_method}
Our main goal is to reduce the computation by removing the Softmax ($exp(.)$) layer. We believe one of the roles of the Softmax layer is to normalize the attention so that each token is a weighted average of the values of all tokens. This ensures that the attention values are bounded. Hence, we introduce an alternative normalization method that does not need a Softmax layer. 

In the regular attention block, if a channel in $Q$ and/or $K$ has large values, that channel may dominate the dot product $QK^T$. This results in other channels being ignored in calculating the attention. We believe this may be one of the reasons leading to superior performance of the multi-head attention (MSA) compared to the single-head one. Since in MSA, the dominating channel can dominate a single head only leaving the other heads still operational. We propose a method to take this solution to the extreme where we can normalize each channel in $Q$ and $K$ across tokens so that different channels become more comparable. We do this by simply dividing the values of each channel by the $\ell_1$ norm of that channel across all tokens:

$$\hat{Q^i} := \frac{Q^i}{|Q^i|_1} \quad \text{and} \quad \hat{K^i} := \frac{K^i}{|K^i|_1}$$

\noindent where $Q^i$ is the $i$'th column of $Q$ (values of the $i$'th channel for all tokens) and $\hat{Q}$ and $\hat{K}$ are the normalized query and key matrices. Given this simple normalization method, we remove the Softmax layer, so the attention block can be written as:

$$O=\hat{Q}\hat{K}^TV$$

\noindent where $O\in \mathbb{R}^{N\times D}$. Similar to standard transformers, we use this block for each head separately, concatenate the outputs, and finally apply the output projection $OW_{proj}$.

One can assume $\hat{Q}\hat{K}^T$ is the attention matrix that quantifies the effect of a token on another token. Interestingly, if the query and key vectors have a large angle, the attention values can become negative, meaning that a token can affect another one negatively. This is in contrast to regular transformers where the attention is always non-negative. A simple pseudo-code is provided in the appendix.

Due to our normalization, the attention values are bounded between $-D$ and $D$. The extremes happen when only a single row of $Q$ and a single row of $K$ are nonzero. In this case, all other tokens will have zero query and key vectors. One may divide the attention by $D$ to bound it between $-1$ and $1$. This constant scalar multiplier can be absorbed into $W_v$, the projection matrix for $V$. 

\textbf{The cost of Softmax:} Both XCA and MSA use Softmax for normalization. Softmax needs running $exp(.)$ which is costly. MSA uses Softmax on a matrix of size $N \times N$ while XCA uses Softmax on a matrix of size $D \times D$. Hence, the order of $exp(.)$ operations is $O(HN^2)$ for MSA and $O(HD^2)$ for XCA. Therefore, Softmax will be bottleneck when increasing the number of tokens (higher image resolutions) in MSA and number of channels (higher capacity transformers) in XCA. On the other hand, our attention block does not use $exp(.)$ operation at all. 
Moreover, in the last row of Table \ref{tab:main_imagenet}, we show that changing GELU to ReLU in SimA gets comparable accuracy to the main experiment (79.6\% vs 79.8\%). This version of SimA does not use any $exp(.)$ operation at the inference time. The reduction in the computation cost of Softmax for a single attention block is shown in Fig.~\ref{fig:softmax_cost}-left for $N>D$ and in Fig.~\ref{fig:softmax_cost}-right for $N<D$ and Table~\ref{tab:exec_time}. This Figure shows the speed-up due to removing Softmax only and does not include the speed-up due to changing the order of multiplications. We believe removing the cost of $exp(.)$ operation can have a large impact particularly in edge devices with limited resources.


\section{Related Work}

\textbf{Vision Transformers:} Convolutional Neural Networks (CNNs) have become ubiquitous as the most commonly used network architecture for computer vision tasks \cite{krizhevsky2012imagenet,he2016deep,chollet2017xception,howard2017mobilenets}. 
Transformers have recently emerged as a promising alternative to CNNs. Transformers \cite{vaswani2017attention} rely entirely on self-attention mechanism and was originally introduced for NLP tasks. ViT \cite{dosovitskiy2020image} adapts transformers to obtain convolution-free architecture for computer vision tasks by dividing each image in $16\times16$ patches and considering each patch as a token input. DeiT \cite{touvron2021deit} improves training efficiency of ViT on smaller dataset. The Scaled Dot-Product Attention module \cite{vaswani2017attention} used by transformers rely on the softmax operation for normalization. Unlike CNNs/MLP \cite{krizhevsky2012imagenet,simonyan2014very,touvron2021resmlp,tolstikhin2021mlp} based architectures, softmax is an important part of transformer architecture. In this paper, we address replacing the softmax ($exp(.)$) operation in the self-attention module of vision transformers.

\textbf{Efficient Vision Transformers:} Transformers have a large memory footprint, so deploying them on edge devices with limited resources is difficult. Many works study efficiency of transformers \cite{yu2022metaformer, brown2022dartformer,keles2022computational,cai2022efficientvit}. LeViT \cite{graham2021levit} uses down-sampling in stages to improve efficiency. \cite{mehta2021mobilevit,Wu_2021_ICCV} integrate convolution in transformer. \cite{liu2021swin,ho2019axial} improve the self-attention efficiency by limiting the attention of each token to subset of tokens. \cite{lu2020twinbert} uses distillation to improve the efficiency of the network. \cite{fayyaz2021ats,rao2021dynamicvit,marin2021token} decrease the number of tokens by token pruning. \cite{liu2022ecoformer} apply quantization on transformers. Although these works limit the computation generally, softmax or $exp(.)$ function is still required to calculate the attention. Our idea is orthogonal to these methods since we can replace attention block in any transformer with our $exp(.)$ free attention block. 

\begin{table*}[t!]
    \begin{center}
    \caption{\textbf{ImageNet classification:} We denote replacing Softmax attention with SimA by X $\rightarrow$ SimA. Softmax column indicates the number of $exp(.)$ operations in the attention block. $N$ is the number of tokens, $D$ is the token dimension, $H$ is the number of heads, $M$ is the local window size, and $R$ is the reduction ratio. We also report ResNet50 RA (with RandAug \cite{cubuk2020randaugment}). Models indicated by * use teacher during training. EfficientNet outperforms our method, but it is a convolutional network and uses more FLOPs at higher image resolution. SOFT also has $exp(.)$ function in the backbone which is costly. Purple rows are our method while blue rows are comparable baselines. Our method is a $exp(.)$ free transformer and has on-par accuracy with SOTA transformers. To simplify SimA even further, we investigate two more variations in yellow rows: (1) Replacing GELU with ReLU, (2) Replacing multi-head attention with single head attention. Interestingly, SimA has comparable performance even with single head attention and ReLU. Note that the ReLU version does not need any $exp(.)$ operation at the inference time. 
    Execution time of SimA and other baselines are shown in Fig.~\ref{fig:softmax_cost} and Table~\ref{tab:exec_time}.
    }
    \label{tab:main_imagenet}
    \scalebox{0.68}
    {
    \begin{tabular}{llccccc} %
    \toprule
    &Model  & params & FLOPs & Resolution   & Softmax/\#$exp$ & Top1-Acc  \\
    \midrule
    CNN & ResNet18 \cite{he2016deep} & \pzo12M & \dzo1.8B   & $224$ & $0$ & 69.8 \\
    \rowcolor{lime}
    Transformer & XCiT-T24/16 \cite{ali2021xcit} & \pzo12M & \dzo2.3B   & $224$ & $HD^2$ & 79.4 \\
    \rowcolor{pink} 
    Transformer & XCiT-T24/16 $\rightarrow$ SimA & \pzo12M & \dzo2.3B   & $224$ & $0$ & 79.8 \\
    \rowcolor{lime}
    Transformer & XCiT-T12/8 \cite{ali2021xcit} & \pzo7M & \dzo4.8B   & $224$ & $HD^2$ & 79.7 \\
    \rowcolor{pink} 
    Transformer &XCiT-T12/8 $\rightarrow$ SimA & \pzo7M & \dzo4.8B   & $224$ & $0$ & 79.4 \\
    \toprule
    \multirow{4}{*}{CNN}&ResNet50 RA \cite{cubuk2020randaugment} & \pzo25M & \dzo3.9B   & $224$ & $0$ & 77.6 \\ 
    &EfficientNet-B5 RA \cite{cubuk2020randaugment} & \pzo30M & \dzo9.9B   & $456$ & $0$ & 83.9 \\ 
    &RegNetY-4GF  \cite{radosavovic2020designing}  & \pzo21M & \pzo4.0B   & $224$  & $0$ & 80.0  \\
    &ConvNeXt-T  \cite{liu2022convnet}  & \pzo29M & \pzo4.5B   & $224$  & $0$ & 82.1  \\
    \midrule
    \multirow{3}{*}{MLP} & ResMLP-S24  \cite{touvron2021resmlp}  & \pzo30M & \pzo6.0B   & $224$  & $0$ & 79.4  \\
    &MS-MLP-T  \cite{zheng2022mixing}  & \pzo28M & \pzo4.9B   & $224$  & $0$ & 82.1  \\
    &Hire-MLP-S  \cite{guo2021hire}  & \pzo33M & \pzo4.2B   & $224$  & $0$ & 82.1  \\
    \midrule
    &Twin-SVT-S  \cite{NEURIPS2021_4e0928de}  & \pzo24M & \pzo3.7B   & $224$  & $HM^2N$ & 81.7  \\
    \rowcolor{lime}
    &CvT-13  \cite{wu2021cvt}  & \pzo20M & \pzo4.5B   & $224$  & $HN^2$ & 81.6  \\
    \rowcolor{pink}
    \multirow{-3}{*}{Hybrid}&CvT-13 $\rightarrow$ SimA  & \pzo20M & \pzo4.5B   & $224$  & $0$ & 81.4  \\
    \midrule
    
    \multirow{11}{*}{Transformer}&Swin-T  \cite{liu2021swin}          & \pzo29M & \dzo4.5B  & $224$  & $HM^2N$ & 81.3  \\
    &PVT-S  \cite{wang2021pyramid}          & \pzo24M & \dzo4.0B  & $224$  & $HN^2/R$ & 79.8  \\
    &T2T-ViT-14  \cite{Yuan_2021_ICCV}          & \pzo21M & \dzo5.2B  & $224$  & $HN^2$ & 80.7  \\
    &CaiT-XS24* \cite{touvron2021going}& \pzo26M & \dzo19.3B  & $384$  & $HN^2$ & 84.1 \\
    &SOFT-S  \cite{lu2021soft}          & \pzo24M & \dzo3.3B  & $224$  & $HN^2$ & 82.2  \\
    
    &DeiT-S*  \cite{touvron2021deit}  & \pzo22M & \dzo4.6B   & $224$  & $HN^2$ & 81.2  \\
    &XCiT-S12/16*\cite{ali2021xcit} & \pzo26M & \dzo4.8B & $224$  & $HD^2$ & 83.3 \\
    \rowcolor{lime}
    &DeiT-S  \cite{touvron2021deit}  & \pzo22M & \dzo4.6B   & $224$  & $HN^2$ & 79.8  \\
    \rowcolor{lime}
    &XCiT-S12/16\cite{ali2021xcit} & \pzo26M & \dzo4.8B & $224$  & $HD^2$ & 82.0 \\
    \rowcolor{pink} 
    &DeiT-S $\rightarrow$ SimA & \pzo22M & \dzo4.6B & $224$  & $0$ & 79.8 \\
    \rowcolor{pink}
    &XCiT-S12/16 $\rightarrow$ SimA & \pzo26M & \dzo4.8B & $224$  & $0$ & 82.1 \\
    \midrule
    \rowcolor{lemon}
    \scriptsize{Multi-Head/GELU} & DeiT-S $\rightarrow$ SimA & \pzo22M & \dzo4.6B & $224$  & $0$ & 79.8 \\
    \rowcolor{lemon}
    \scriptsize{Multi-Head $\rightarrow$ Single-Head} & DeiT-S $\rightarrow$ SimA & \pzo22M & \dzo4.6B & $224$  & $0$ & 79.4 \\
    \rowcolor{lemon}
    \scriptsize{GELU $\rightarrow$ ReLU} & DeiT-S $\rightarrow$ SimA & \pzo22M & \dzo4.6B & $224$  & $0$ & 79.6 \\

    \bottomrule
    
    \end{tabular}
    }
    \end{center}
    \vspace{-.15in}
\end{table*}

\textbf{Linear Attention:} Vanilla attention has $O(N^2D)$ computation and memory complexity, where $N$ is number of input tokens and $D$ is dimension of each token. 
Some works target this issue by replacing vanilla attention with a linear attention with $O(ND^2)$ complexity. XCiT  \cite{shen2021efficient,ali2021xcit} uses attention across feature channels rather than tokens. Some works use similarity kernels to approximate softmax, thus it is possible to have linear complexity by doing $\phi(Q)(\phi(K)^T\phi(V))$ instead of $(\phi(Q)\phi(K)^T)\phi(V)$ where $\phi(x)$ is the kernel function. \cite{katharopoulos2020transformers} uses $\phi(x) = 1+elu(x)$, whereas \cite{lu2021soft,peng2021random} use Gaussian kernel functions. \cite{xiong2021nystromformer} use SVD decomposition and \cite{choromanski2020rethinking} use positive random features to approximate softmax. \cite{wang2020linformer} approximate attention with a low rank matrix. All these methods either use exponential function. For example, SOFT \cite{lu2021soft} removes Softmax without reducing the number of $exp(.)$ operations. Our ideas are different since we aim to remove the costly $exp(.)$ operation. Moreover, the focus of those methods is to have linear attention with respect to number of tokens which is not the main focus of this paper. A recent work in the NLP community, CosFormer \cite{qin2022cosformer}, passes $Q$ and $K$ through a ReLU unit and normalizes their product. It also adds a re-weighting method that improves the locality of the data using $sine(.)$ and $cosine(.)$ functions. Our idea is simpler and we apply it to visual recognition rather than NLP. Moreover, cosine re-weighting in CosFormer requires $4\times$ more FLOPs in $K$ and $V$ dot product compared to ours.

\textbf{Softmax Approximation:} Softmax is an expensive operation on hardware since it requires $exp(.)$ operation. More specifically, softmax in transformer architecture contributes to major part of computation when the input is large \cite{stevens2021softermax}. 
\cite{banerjee2020exploring} approximates softmax with Taylor expansions, whereas \cite{gao2020design,du2019efficient,ham20203,zhu2020efficient} target designing a hardware architecture to approximate softmax. Softermax \cite{stevens2021softermax} uses a low-precision implementation of $2^x$. \cite{zafrir2019q8bert} uses lower precision computation. \cite{prato2019fully,lin2020towards} use quantized softmax. While these works approximate Softmax at the hardware, we replace Softmax completely with $\ell_1$ normalization at the model architecture.

\section{Experiments}
We evaluate effectiveness of SimA attention block by replacing self-attention in three popular vision transformer families: DeiT, XCiT and CvT. We evaluate our model on image classification, object detection, image segmentation, and self-supervised learning.

\subsection{Image Classification}
\label{ImageNet}

\begin{table}[t!]
    \begin{center}
    \caption{\textbf{Linear Attention Comparison:} We compare SimA with previous linear attention methods introduced in NLP. We report ImageNet Top-1 validation accuracy. Note that the focus of these methods is to have linear attention with respect to the number of tokens while the main focus of SimA is to remove $exp(.)$ operation. * 
    CosFormer is originally in NLP. We ran multiple versions of CosFormer with cosine re-weighting (multiple learning rates and weight decays) for the vision task, however, none of them converged. Moreover, CosFormer with cosine re-weighting requires $4\times$ more FLOPs compared to SimA in multiplying $Q$, $K$, and $V$ matrices. More details are in the appendix. Execution time  of SimA and SOFT is shown in Fig.~\ref{fig:softmax_cost} and Table \ref{tab:exec_time}.
    }
    \label{tab:linear_attention}
    \scalebox{0.7}
    {
    \begin{tabular}{lcccc} %
    \toprule
    Model  & params & FLOPs & Softmax/\#$exp$ & Top1-Acc \\
    \midrule
    Transformer \cite{vaswani2017attention} & \pzo13M & \dzo3.9B  & $HN^2$ & 79.1 \\
    Linformer \cite{wang2020linformer} & \pzo13M & \dzo1.9B  & $HN$ & 78.2 \\
    Performer \cite{choromanski2020rethinking} & \pzo13M & \dzo2.2B  & $ND$ & 76.1 \\
    Nyströmformer \cite{xiong2021nystromformer} & \pzo13M & \dzo2.0B  & $HN$ & 78.6 \\ 
    SOFT \cite{lu2021soft} & \pzo13M & \dzo1.9B  & $HN^2$ & 79.3 \\ 
    \rowcolor{pink}
    XCiT-T20/16 $\rightarrow$ SimA & \pzo12M & \dzo1.9B   & $0$ & 79.2 \\
    \midrule
    XCiT w/ Efficient Attention \cite{shen2021efficient} & \pzo22M & \dzo4.8B   & $ND$ & 80.9 \\
    CosFormer w/o re-weighting * \cite{qin2022cosformer} & \pzo22M & \dzo4.8B   & $0$ & 76.1 \\
    \rowcolor{pink}
    XCiT-S12/16 $\rightarrow$ SimA & \pzo22M & \dzo4.8B   & $0$ & 82.1 \\

    \bottomrule
    
    \end{tabular}
    }
    \end{center}
    \vspace{-.2in}
\end{table}

\textbf{Dataset:} We train on ImageNet1K \cite{deng2009imagenet} and report Top-1 accuracy on the validation set.

\textbf{Implementation Details:} We use PyTorch \cite{paszke2019pytorch} and Timm \cite{wightman2019pytorch} libraries to train our models with a setup similar to \cite{ali2021xcit,touvron2021deit}. We use AdamW \cite{loshchilov2017decoupled} optimizer. We train CvT and DeiT models with $300$ epochs and XCiT models with $400$ epochs. We set the batch size to $1024$ and weight decay to $0.05$. We use cosine scheduling with an initial learning rate of $5e-4$. We use Stochastic depth drop rate \cite{huang2016deep} of $0.05$. Data augmentations are the same as those in \cite{touvron2021deit} including Rand-Augment \cite{cubuk2020randaugment}, CutMix \cite{yun2019cutmix} and Mixup \cite{zhang2017mixup}. Following \cite{ali2021xcit,touvron2021going}, we train our models with images of resolution $224$ and evaluate it using images with a crop ratio of 1.0. Training DeiT-S or XCiT-S12/16 with $8$ RTX $6000$ GPUs takes approximately 100 hours.  

We use SimA along with the following three transformer architectures to show its generalization:


\textbf{- DeiT}: \cite{touvron2021deit} is a well-known transformer architecture based on ViT \cite{dosovitskiy2020image}. Since we do not use the distillation token introduced in DeiT, in our setting, DeiT is very similar to ViT except that it converged faster due to better optimization parameters. We use DeiT-S which has the following settings: patch size$=16$, embedding dimensions$=384$, number of heads$=6$ and layers$=12$. Self-attention in DeiT has complexity of $O(DN^2)$ which is quadratic on the number of tokens $N$. 


\textbf{- XCiT}: \cite{ali2021xcit} is a state-of-the-art vision transformer architecture with a linear attention. XCiT has 2 major differences compared to DeiT: \begin{enumerate*}[label={(\arabic*)}]
    \item XCiT has Local Patch Interaction (LPI) in each block, which consists of one depth-wise 3×3 convolution followed by Batch Normalization, GELU and another depth-wise 3×3 convolution. 
    \item XCiT has separate class attention layers similar to \cite{touvron2021going}. The \texttt{CLS} token is added at the end of the initial self-attention stage and class attention layers are used to aggregate information from image tokens to the class token. This modification adds extra parameters and computation to the model.

\end{enumerate*}

\begin{table}[t!]
    \begin{center}
    \caption{\textbf{Execution Time Comparison:} We compare execution time of different attention blocks on 3 different edge devices. SimA has faster execution time in edge devices due to removing $exp(.)$ operation. To measure the effect of $exp(.)$ only, we fix the order of ($QK^TV$) product so that all models have the same dot product complexity. We set $N>D$ for top table and $N<D$ for the bottom table. These results are also shown in Figure \ref{fig:softmax_cost}.
    }
    \label{tab:exec_time}
    \scalebox{0.68}
    {
    \begin{tabular}{|l|ccc|} %
    \toprule
    \multicolumn{4}{|c|}{$D=64 \quad N=256 \quad H=8 \quad N>D$}\\\midrule
    Model &\multicolumn{3}{c|}{Execution Time (ms)}\\\midrule
    & NVIDIA nano & Apple M1 & Raspberry Pi 4 \\
    \midrule
    Vanilla Attention (DeiT) & 177.4 & 35.8 & 180.6\\
    ELU \cite{katharopoulos2020transformers} &	144.5 &	26.2 & 160.0 \\
    \rowcolor{pink}
    SimA & \textbf{63.6}	& \textbf{20.4} & \textbf{51.9}  \\
    \midrule\midrule
    \multicolumn{4}{|c|}{$D=256 \quad N=64 \quad H=8 \quad N<D$}\\\midrule
    Model &\multicolumn{3}{c|}{Execution Time (ms)}\\\midrule
     & NVIDIA nano & Apple M1 & Raspberry Pi 4 \\
     \midrule
     SOFT \cite{lu2021soft} & 145.9 & 36.0 & 177.4\\
     XCA \cite{shen2021efficient} & 144.4 & 36.1 & 178.7\\
     ELU \cite{katharopoulos2020transformers} & 105.2 & 26.4 & 160.1\\
    \rowcolor{pink}
     SimA & \textbf{58.0} & \textbf{20.5} & \textbf{53.2}\\
    
    \bottomrule
    
    \end{tabular}
    }
    \end{center}
    \vspace{-.2in}
\end{table}

We replace SimA in three variant of XCiT: XCiT-S12/16, XCiT-T12/8 and XCiT-T24/16. XCiT-S12/16 has a patch size of $16$, embedding dimension of $384$, $8$ heads, $12$ layers, and $2$ class attention layers. XCiT-T12/8 is similar to XCiT-S12/16 with a patch size of $8$, embedding dimension of $192$, and $4$ heads. XCiT-T24/16 is similar to XCiT-T12/8 with patch size of $16$.

\textbf{- CvT}: We apply SimA to CvT \cite{wu2021cvt}, which is a SOTA hybrid convolution/transformer architecture. CvT has 3 stages. Each stage has a Convolution Token Embedding layer followed by transformer blocks. We use CvT-13 in our experiments which $13$ blocks in total. 

\textbf{Results on ImageNet:} 
We replace MSA and XCA blocks with our SimA block in DeiT, CvT and XCiT respectively, and train our models on ImageNet. Note that we train our models from scratch without distillation from a teacher. Results are in Table \ref{tab:main_imagenet}. In XCiT models, we get comparable results when replacing XCA block with SimA block. Compared to DeiT-S, our attention block performs on-par with DeiT-S. Moreover, our method with no Softmax layer, achieves comparable accuracy ($0.2$ point lower) compared to CvT-13. This suggests that one can replace attention block with SimA in these standard SOTA transformers without degrading their performance. Since SimA is $exp(.)$ free, it has the advantage over regular attention architectures in terms of efficiency and simplicity.  

\textbf{Comparison to Linear Attention:} We compare SimA with other Linear Attention methods in NLP literature in Table \ref{tab:linear_attention}. We train all methods with ImageNet-1K training set and report Top-1 accuracy on ImageNet-1K validation set. SimA has better or on-par accuracy compared to other methods. Additionally, SimA is $exp(.)$ free which is the main goal of this work. 

\begin{table*}[t!]
    \begin{center}
    \caption{\textbf{Transfer to MS-COCO dataset:} Models with * are pretrained with a teacher on ImageNet. Swin-T has more parameters and Softmax overhead. XCiT-S12/8 has $4\times$ more tokens. Our method is $exp(.)$ free, thus it is more efficient for high resolution images and high capacity models (Fig.~\ref{fig:softmax_cost}). Execution time of SimA and XCA (XCiT) is shown in Fig.~\ref{fig:softmax_cost}.}
    \label{tab:transfer_coco}
    \scalebox{0.80}{
    \begin{tabular}{lcc|ccc|ccc}
        \toprule
             \multicolumn{3}{c}{} &\multicolumn{3}{c}{Detection} & \multicolumn{3}{c}{Segmentation}  \\
             \midrule
             Backbone & params & $exp(.)$ & $\text{AP}^{box}$ & $\text{AP}^{box}_{50}$ & $\text{AP}^{box}_{75}$ & $\text{AP}^{mask}$ & $\text{AP}^{mask}_{50}$ & $\text{AP}^{mask}_{75}$\\ 
            
            \midrule
            ResNet50 \cite{he2016deep} & 44.2M & \ding{55} & 41.0 & 61.7 & 44.9 & 37.1 & 58.4 & 40.1 \\
            PVT-Small \cite{wang2021pyramid} & 44.1M & \ding{51} & 43.0 & 65.3 & 46.9 & 39.9 & 62.5 & 42.8 \\
            ViL-Small \cite{zhang2021multi} & 45.0M & \ding{51} & 43.4 & 64.9 & 47.0 & 39.6 & 62.1 & 42.4 \\
            Swin-T \cite{liu2021swin} & 47.8M & \ding{51} & 46.0 & 68.1  & 50.3 & 41.6 & 65.1 & 44.9 \\
            
            XCiT-S12/16* & 44.3M & \ding{51} & 45.3 & 67.0 & 49.5 & 40.8 & 64.0 & 43.8  \\
            
            XCiT-S12/8*  & 43.1M & \ding{51} & 47.0 & 68.9  & 51.7  & 42.3  & 66.0 & 45.4  \\
            \rowcolor{lime}
            XCiT-S12/16  & 44.3M & \ding{51} & 45.0 & 66.7  & 48.9  & 40.5  & 63.6 & 43.2  \\
            \rowcolor{pink} 
            XCiT-S12/16 $\rightarrow$ SimA & 44.3M & \ding{55} & 44.8 & 66.5  & 48.8  & 40.3  & 63.2 & 43.3  \\

        \bottomrule     
     \end{tabular}
     }
    \end{center}
    \vspace{-.2in}
\end{table*}

\begin{table*}[t!]
    \begin{center}
    \caption{\textbf{Self-Supervised Learning:} We train SimA attention block with DINO (SSL). Our method achieves performance comparable to transformer models with Softmax and trained for $100$ epochs. Note that methods with different SSL task and higher number of epochs are not directly comparable. Execution time of SimA and XCA (XCiT) is shown in Fig.~\ref{fig:softmax_cost}. }
    \label{tab:ssl}
    \scalebox{0.80}{
    \begin{tabular}{llcccccc}
        \toprule
        SSL Method & Model  & params & epochs & $exp(.)$ &  FLOPs & Linear & $k$-NN \\
        \toprule
        ISD \cite{tejankar2020isd} & ResNet50  & 25M & 200 & \ding{55} & 3.9B & 69.8 & 62.0 \\
        MoCo v2 \cite{he2020momentum} & ResNet50  & 25M & 200 & \ding{55} & 3.9B & 69.9 & - \\
        MSF \cite{koohpayegani2021mean} & ResNet50  & 25M & 200 & \ding{55} & 3.9B & 72.4 & 64.9  \\
        BYOL \cite{grill2020bootstrap} & ResNet50  & 25M & 1000 & \ding{55} & 3.9B & 74.3 & 66.9 \\

        MoBY~\cite{xie2021self} & Swin-T & 29M & 300 & \ding{51}  & \dzo4.5B & 75.0 & -- \\
        \midrule
        DINO~\cite{caron2021emerging} & ResNet-50  & 23M & 300 & \ding{55} &  \dzo4.1B & 74.5 & 65.6 \\
        DINO~\cite{caron2021emerging} & ResMLP-S24  & 30M & 300 & \ding{55} &  \dzo6.0B & 72.8 & 69.4 \\
        DINO~\cite{caron2021emerging} & ViT-S/16  & 22M & 300 & \ding{51} & \dzo4.6B & 76.1 & 72.8 \\
        
        DINO~\cite{caron2021emerging} & XCiT-S12/16 & 26M & 300 & \ding{51} & \dzo4.9B & 77.8 & 76.0 \\
        
        \midrule
        DINO~\cite{caron2021emerging} & ViT-S/16 & 22M & 100 & \ding{51} & \dzo4.6B & 74.0 & 69.3 \\
        \rowcolor{lime}
        DINO~\cite{caron2021emerging} & XCiT-S12/16 & 26M & 100 & \ding{51} & \dzo4.9B & 75.8 & 71.6 \\
        \rowcolor{pink}
        DINO~\cite{caron2021emerging} & XCiT-S12/16 $\rightarrow$ SimA & 26M & 100 & \ding{55} & \dzo4.9B & 75.5 & 71.2 \\

        \bottomrule
        \end{tabular}
        }
    \end{center}
    \vspace{-.2in}
\end{table*}

\subsection{Transfer To Object Detection and Semantic Segmentation}
\label{Transfer_coco}
As shown in Figure \ref{fig:softmax_cost}, Table \ref{tab:exec_time}, and \cite{stevens2021softermax}, softmax operation represents a large fraction of runtime in vision transformers, especially when the image resolution is high. In object detection and segmentation tasks we usually forward high resolution images. 
We demonstrate the transferability of SimA to these dense prediction tasks by fine-tuning our ImageNet pretrained model on them.

\textbf{Dataset:} We use MS-COCO \cite{lin2014microsoft} dataset for these tasks. MS-COCO has $118$K training images and $5$K validation images with $80$ categories. Images are annotated with bounding boxes and semantic segmentation masks. 

\textbf{Implementation Details:} We follow \cite{ali2021xcit,liu2021swin,chen2019mmdetection} for the setup and implementation. We use our pretrained model as the backbone of Mask RCNN \cite{he2017mask}. Similar to \cite{ali2021xcit}, we use FPN \cite{lin2017feature} to extract features from layers  $4$, $6$, $8$ and $12$ of the transformer. We use AdamW \cite{loshchilov2017decoupled} optimizer with a learning rate of $1e-4$ and weight decay $0.05$. We train our model for $36$ epochs with batch size of $16$ on $8$ RTX$2080$Ti GPUs. Training takes 36 hours. 

\textbf{Results on MS-COCO:} We compare our XCiT-S12/16 $\rightarrow$ SimA model with other vision transformers and ResNet in Table~\ref{tab:transfer_coco}. We report the performance on the minival set. For a  fair comparison, we limit the comparison to all models which are initialized with ImageNet1K pretrained backbones and trained with the same training time budget (3x schedule) on MS-COCO dataset. In comparison to other transformers, our method gets on-par performance while it is free of Softmax overhead on high resolution images or high capacity models (refer to Fig.~\ref{fig:softmax_cost}).

\subsection{Self-Supervised Learning}
\label{SSL}

To show the generalizability of SimA, we train our SimA model on a pretext task for self-supervised learning (SSL). We use the non-contrastive task called DINO \cite{caron2021emerging} for SSL pre-training. We train our model on ImageNet train set ($1.2M$) without the use of ground-truth labels. DINO training is relatively expensive since it requires forwarding multi-crop augmentation through two models. Due to limited resources, we train our model and the baselines for $100$ epochs. To train our XCiT-S12/16 $\rightarrow$ SimA model with DINO, we follow the training configuration of XCiT-S12/16 from the official repository of DINO \cite{official_dino_code}. Similar to DINO, we use AdamW optimizer in PyTorch library with initial learning rate of $0.00025$ with cosine scheduling. We use initial weight decay of $0.04$ and increase it to $0.4$ with cosine scheduling. We train for $100$ epochs with minibatches of size $256$. The training takes approximately $100$ hours on four RTX-$3090$ GPUs. We use similar settings for training our method and the baseline (XCiT-S12/16).   

\textbf{Results of SSL training:} Following \cite{caron2021emerging,abbasi2020compress}, we report $k$-NN and Linear evaluation metrics for evaluating the SSL models. For $k$-NN evaluation, we forward images of training and validation set through the frozen backbone and extract features. We report $20$-NN on the validation set. For Linear evaluation, we freeze the backbone and train a linear layer on extracted features from the frozen backbone and report Top-1 accuracy on the ImageNet validation set. We adopt a similar approach to DINO \cite{caron2021emerging} for extracting features from XCiT architecture. We extract the classification tokens of the last two class attention layers and global average pooling of the last two regular attention layers. Each of those 4 vectors is of size $384$. We concatenate them and train a linear layer of size $4 \times 384$ to $1000$ classes of ImageNet1K. We use similar training settings as DINO to train a linear layer for both our method and the baseline (XCiT-S12/16). We train for $100$ epochs with SGD optimizer and the following settings: learning rate: $0.001$ with cosine scheduling, batch size: $1024$, and weight decay: $0$. Results are shown in Table~\ref{tab:ssl}. Our $exp(.)$ free method performs comparably with the baselines with $100$ epochs of training.

\subsection{Single-head vs Multi-head Attention} 
In the regular attention block, if a channel in $Q$ and/or $K$ has large values, that channel may dominate the dot product $QK^T$. We believe multi-head attention (MSA) mitigates this issue to some degree by containing the dominant channel in one head only so that the other heads can have reasonable effect in the final attention. In SimA, by doing $\ell_1$ normalization of each channel in $Q$ and $K$ across tokens, different channels become more comparable in the dot product $QK^T$, so multi-head attention may not have a large effect anymore. To evaluate our hypothesis empirically, we train both DeiT-S $\rightarrow$ SimA and DeiT-S with single head attention only. Results are in Table.~\ref{tab:single_head}. Interestingly, we show that with single-head attention, our method gets comparable results ($0.4$ point lower) while the accuracy of DeiT-S drops by $2.8$ points. This suggests that unlike the vanilla attention block, multi-head attention is not critically important in SimA, which leads to simplicity SimA even further. 

\begin{table}[h!]
    \begin{center}
    \caption{\textbf{Effect of Removing Multi-Head Attention:} In single head variation, our method degrades much less compared to DeiT probably due to normalization of $Q$ and $K$.}
    \label{tab:single_head}
    \scalebox{0.70}{
    \begin{tabular}{l|cc|cc}
    \toprule
    Model & \multicolumn{2}{c|}{DeiT-S $\rightarrow$ SimA} & \multicolumn{2}{c}{DeiT-S} \\
    Attention Heads & $6$ (Multi-Head) & $1$ (Single) & $6$ (Multi-Head) & $1$ (Single) \\
    \midrule
    ImageNet Top-1 acc. & 79.8 & 79.4 (\textcolor{magenta}{-0.4}) & 79.8 & 77.0 (\textcolor{magenta}{-2.8})\\
    
    \bottomrule
    \end{tabular}
    }
    \end{center}
    \vspace{-.2in}
\end{table}

\subsection{Replacing GELU with ReLU}
\vspace{-.05in}
Similar to Softmax function, GELU activation function also uses $exp(.)$ operation, which is costly. We replace all GELU activation functions in DeiT-S $\rightarrow$ SimA with ReLU. We observe that DeiT-S $\rightarrow$ SimA with ReLU gets accuracy of $79.6$ which is only $0.2$ points lower than DeiT-S $\rightarrow$ SimA with GELU activation function. Note that SimA with ReLU does not use any $exp(.)$ operation at the inference time, leading to further efficiency of the model. Results are in Table \ref{tab:main_imagenet} (yellow rows).

\subsection{Effect of $\ell_1$ Normalization}
\label{ablation}
\vspace{-.05in}
To see the effect of $\ell_1$ normalization, we train our model without normalizing $Q$ and $K$. We use XCiT-S12/16 $\rightarrow$ SimA with the same hyperparameters as our main experiment in Section \ref{ImageNet}. Note that without normalization, the range of $QK^T$ can be from $-\infty$ to $+\infty$. None of our several trials converged as the training becomes unstable and results in a frequent NaN loss. Moreover, we replace $\ell_1$ with $\ell_2$ normalization, results in 2.9 points drop in accuracy ($82.1\%$ vs $79.2\%$).

\begin{figure}[t]
\centering
\includegraphics[width=0.8\linewidth]{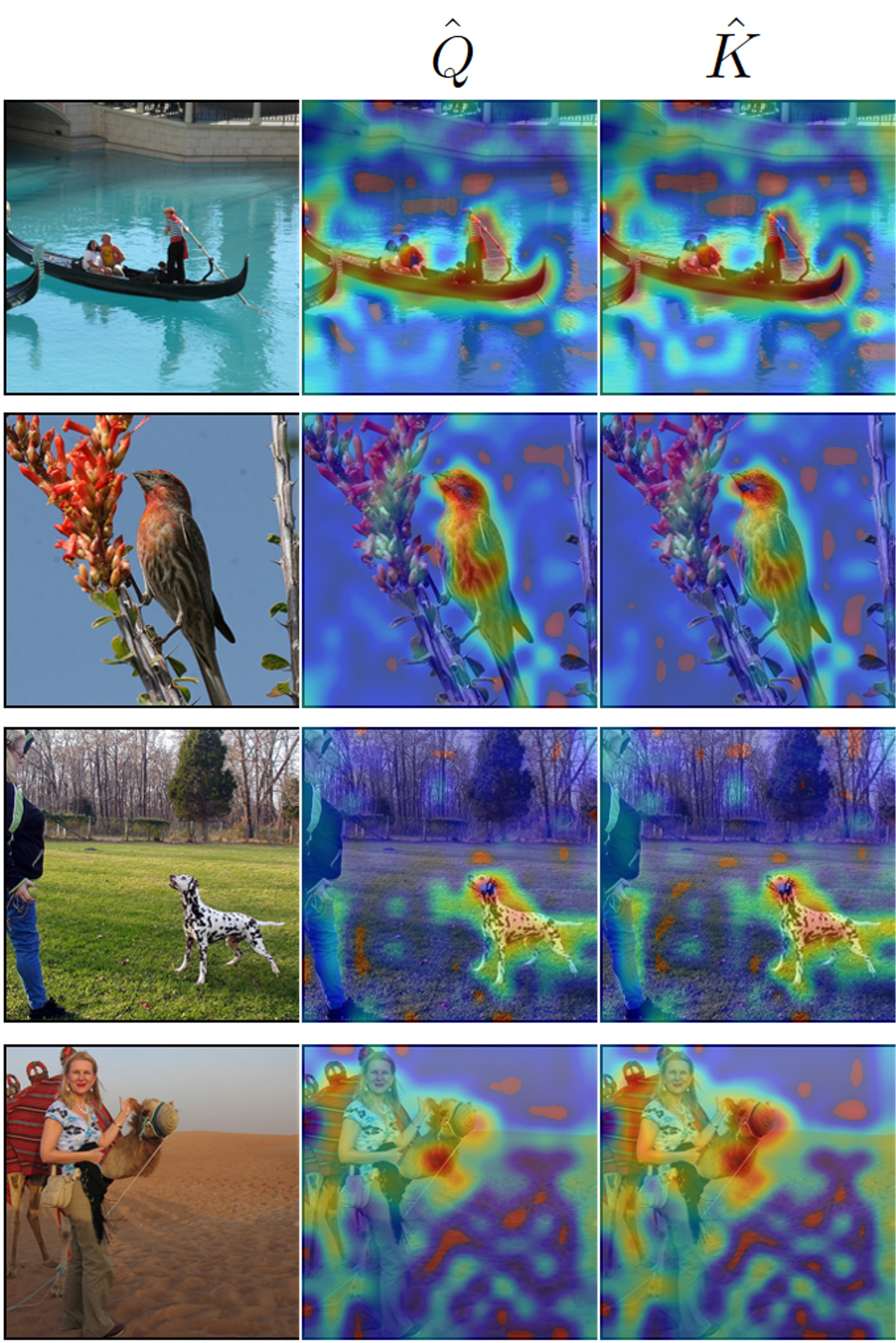}
\vspace{-.5em}

\caption{{\bf Our method (SimA):} Standard attention passes $QK^T$ through Softmax before multiplying with $V$. However, we multiply $\hat{Q}\hat{K}^T$ directly with $V$. Hence, in our case, the magnitude of $\hat{Q}\hat{K}^T$ should identify which tokens are more important (their information flows to the next layers). We show that this magnitude is correlated with the importance of tokens. We extract $\hat{Q}$ and $\hat{K}$ from layer $12$ of transformer. We get $\ell_2$-norm of each token for $\hat{Q}$ and $\hat{K}$, normalize it to range [0,1] and overlay it as a heatmap on the image. We show the same visualization for DeiT in the supplementary for completeness.
We provide more examples in the appendix.}
\label{fig:vis1}
\vspace{-.2in}
\end{figure}

\subsection{Visualization}
\vspace{-.05in}
\label{text_visualization}
Since in SimA, we multiply the dot product $\hat{Q}\hat{K}^T$ directly with $V$ without any Softmax layer, the $\hat{Q}$ and $\hat{K}$ with larger magnitute will have more effect in the output of the block. Hence, we believe this magnitude can highlight the important tokens or image regions. This can be seen as a form of explanation or saliency map. First, we extract $\hat{Q}$ and $\hat{K}$ in the last layer of transformer (layer $12$). Then, we calculate the $\ell_2$-norm of $\hat{Q}$ along the channel dimension to get a single non-negative scalar for each token. We reshape this $N\times1$ vector to the image shape, up-sample it to original image size, normalize it to range $[0,1]$, and overlay it on the image as a heatmap. We repeat the same for $\hat{K}$. As shown qualitatively in Fig.~\ref{fig:vis1}, such a visualization highlights the important regions of the image. 


\section{Conclusion}

We introduced SimA, a simple attention block that does not involve $exp(.)$ operation, to reduce the computational cost of transformers particularly at edge devices. SimA performs normalization on key and query matrices before multiplying them, enabling dynamically switching between $O(DN^2)$ or $O(D^2N)$ depending on the number of tokens (e.g., image resolution). Our extensive experiments show that while reducing the cost of inference, SimA achieves on-par results compared to SOTA methods on various benchmarks including ImageNet classification, MS-COCO object detection and segmentation, and self-supervised learning. Moreover, a single-head variation of SimA, which is even simpler, achieves results on-par with SOTA multi-head attention models. We believe SimA can encourage research in this direction leading to easier adoption of transformers on edge devices with limited resources. \\

\textbf{Acknowledgments:}
This work is partially funded by NSF grant 1845216 and DARPA Contract No. HR00112190135. Any opinions, findings, and conclusions or recommendations expressed in this material are those of the authors and do not necessarily reflect the views of the funding agencies.
Moreover, we would like to thank K L Navaneet, Vipin Pillai, Kossar Pourahmadi for the valuable discussions and proof-reading the paper.

{\small
\bibliographystyle{ieee_fullname}
\bibliography{egbib}
}

\renewcommand{\thefigure}{A\arabic{figure}}
\renewcommand{\thetable}{A\arabic{table}}
\setcounter{figure}{0}
\setcounter{table}{0}

\newpage
\appendix
\section{Appendix}

\subsection{Inference Time Comparison on GPU:} 
\begin{figure*}[t]
    \centering
    \begin{minipage}{0.5\textwidth}
        \centering
        \includegraphics[width=1.0\linewidth]{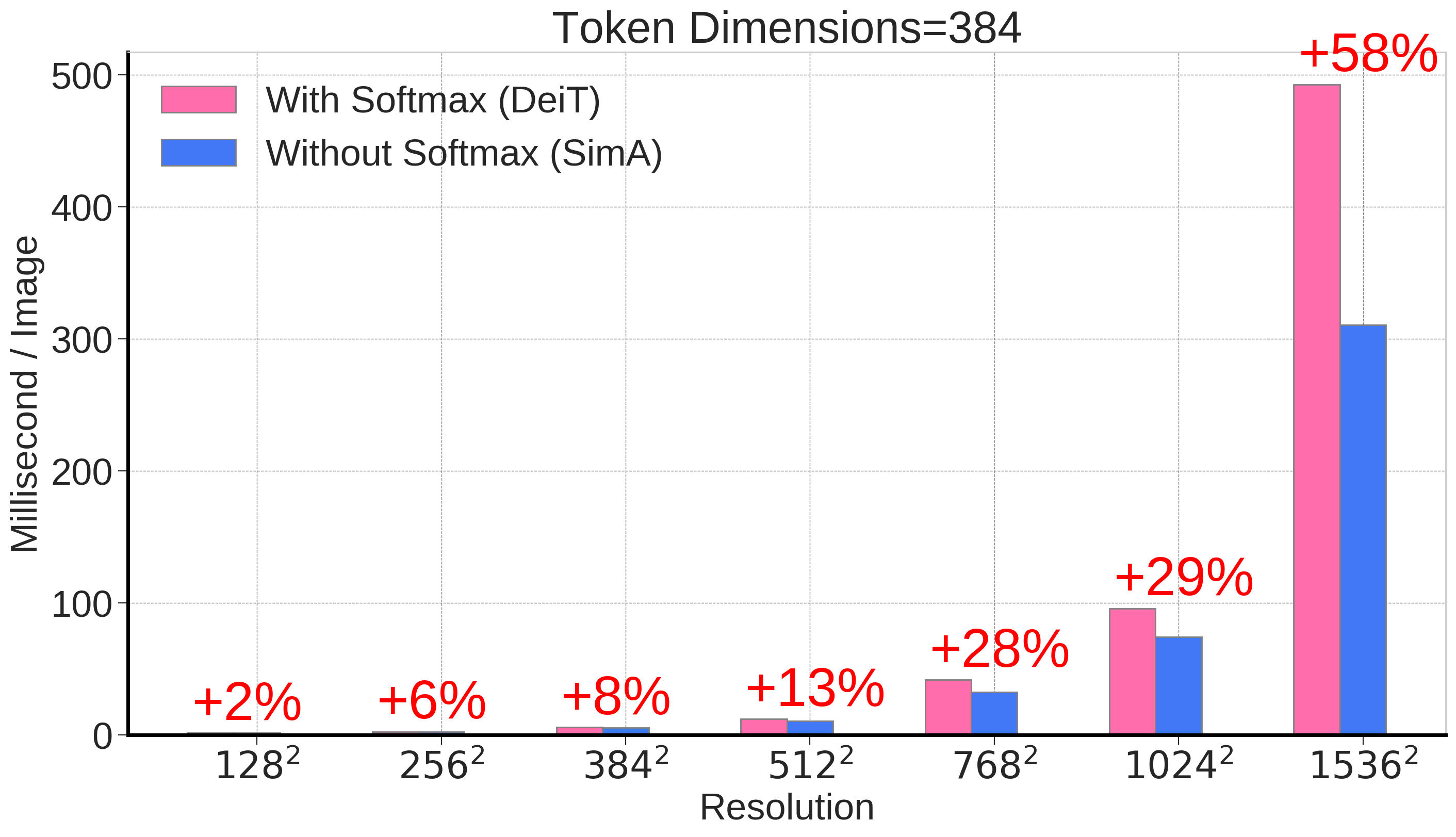}
    \end{minipage}%
    \begin{minipage}{0.5\textwidth}
        \centering
        \includegraphics[width=1.0\linewidth]{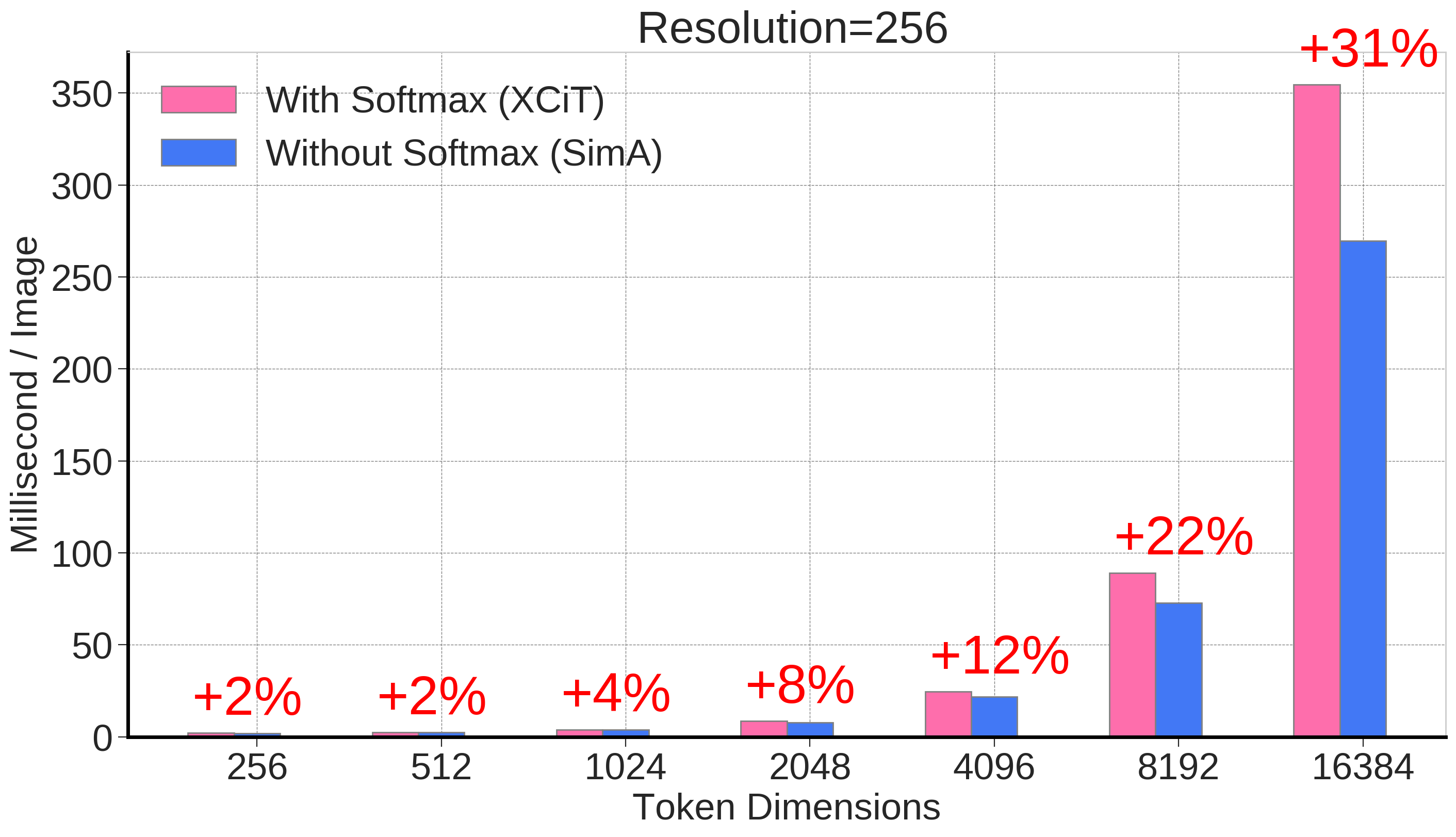}
    \end{minipage}
    \caption{\textbf{Effect of Softmax on inference time (GPU):} We evaluate performance of each model on a single RTX 8000 GPU with batch size of $8$. When comparing the baseline to our method (SimA), we fix the order of ($QK^TV$) to have the same dot product complexity as the baseline. For example, when comparing with DeiT, if $N>D$, then it is more efficient to do $\hat{Q}(\hat{K}^TV)$ for our method, but we do $(\hat{Q}\hat{K}^T)V$ to have same complexity as DeiT($O(N^2D)$). We do this to solely evaluate the effect of Softmax on the computation time. {\bf Left:} We fix the token dimension to $384$ and increase the image resolution. At $1536\times 1536$ resolution, DeiT is $58\%$ slower than our method due to the overhead of $exp(.)$ function in Softmax. {\bf Right:} We fix the resolution and increase the capacity of the model (dimensions of $Q$ and $K$). With $8192$ dimensions, XCiT is $22\%$ slower due to Softmax overhead. } 
    \label{fig:softmax_cost_supp}
    \vspace{-.1in}
\end{figure*}

We compare execution time of SimA and other SOTA methods on edge devices in Figure \ref{fig:softmax_cost}. Additionally, we compare execution time of SimA, XCiT, and DeiT on GPU in Figure \ref{fig:softmax_cost_supp}.  

\subsection{Simple Pseudocode of SimA:}

Since our method is simple, we include the pseudocode of SimA in Algorithm \ref{alg:code}.  

\begin{algorithm*}[ht]
\caption{Pseudocode of SimA (Single Head) in a PyTorch-like style.}
\label{alg:code}
\definecolor{codeblue}{rgb}{0.9,0.25,0.6}
\lstset{
  backgroundcolor=\color{white},
  basicstyle=\fontsize{7.2pt}{7.2pt}\ttfamily\selectfont,
  columns=fullflexible,
  breaklines=true,
  captionpos=b,
  commentstyle=\fontsize{7.2pt}{7.2pt}\color{codeblue},
  keywordstyle=\fontsize{7.2pt}{7.2pt},
}
\begin{lstlisting}[language=python]
# self.qkv: nn.Linear(dim, dim * 3, bias=qkv_bias) ; query, key, value projection
# self.proj: nn.Linear(dim, dim, bias=output_proj_bias) ; output projection 

def forward(self, x):
    B, N, D = x.shape # B: batch size, N: number of Tokens, D: Dimension of Tokens
    qkv = self.qkv(x).reshape(B, N, 3, D).permute(2, 0, 1, 3) # (3 x B x N x D)
    q, k, v = qkv[0], qkv[1], qkv[2]   # split into query (B x N x D), key (B x N x D) and value (B x N x D)

    k = torch.nn.functional.normalize(k, p=1.0, dim=-2) # Normalized key (B x N x D)
    q = torch.nn.functional.normalize(q, p=1.0, dim=-2) # Normalized query (B x N x D)
    
    if (N/D) < 1:
        x = (q @ k.transpose(-2, -1)) @ v # (B x N x D)
    else:
        x = q @ (k.transpose(-2, -1) @ v) # (B x N x D)

    x = self.proj(x) # Output (B x N x D)
    return x
\end{lstlisting}
\end{algorithm*}

\begin{figure*}[h!]
\centering
\vspace{-0.5em}
\includegraphics[width=0.82\linewidth]{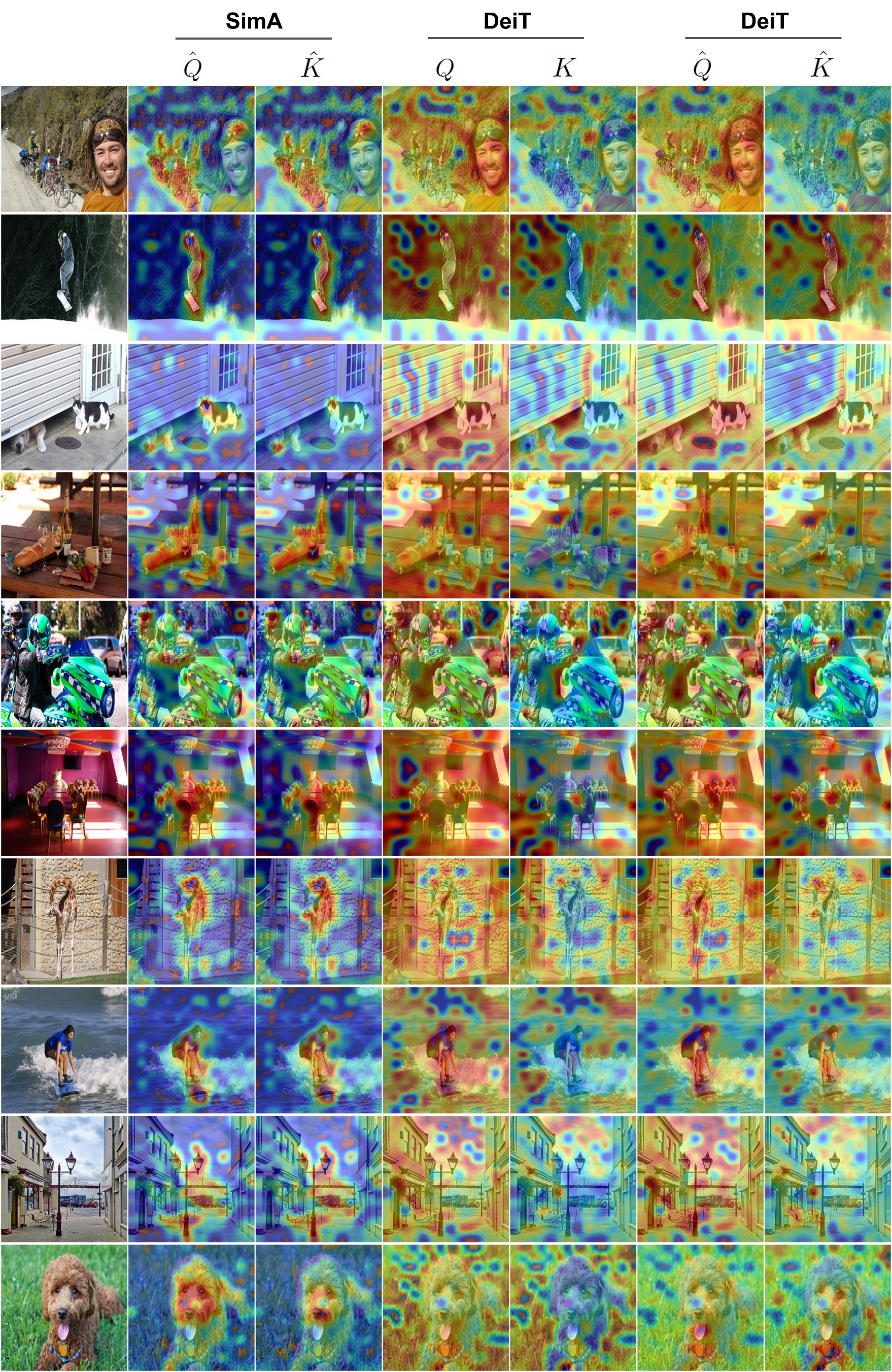}
\vspace{-.5em}

\caption{{\bf Our method (SimA):} We extract $\hat{Q}$ and $\hat{K}$ from layer $12$ of transformer. We get $\ell_2$-norm of each token for $\hat{Q}$ and $\hat{K}$, normalize it to range [0,1] and overlay it as a heatmap on the image. Interestingly, magnitude of tokens represent the significance of tokens in our method. Note that all images are randomly selected from MS-COCO test set without any visual inspection or cherry picking.}
\label{fig:vis2_supp}
\vspace{-.2in}
\end{figure*}

\textbf{Visualization:}
Figure \ref{fig:vis2_supp} provides more results similar to Figure \ref{fig:vis1}. Please see Section 4.7 for details.

\subsection{SimA without LPI:}

Although XCiT \cite{ali2021xcit} shows that LPI layer can improve the accuracy by $1.2$ point, it limits the application of vanilla transformer (e.g., running masked auto encoder models like MAE \cite{he2021masked} is not straightforward). 
To show that our method is not dependent on LPI, we train our model without LPI. We observe that the accuracy drops by $1.2$ point ($82.1\%$ vs $80.9\%$). Hence, although LPI boosts the accuracy, our method has comparable performance without LPI.

\subsection{Details of Linear Attention Comparison:} 
CosFormer with cosine re-weighting requires $4\times$ more FLOPs compared to SimA in multiplying $K$ and $V$ matrices. Since CosFormer is developed for NLP, it assumes one dimensional indexing for the tokens. However, applying it to vision, we need to index the tokens with two indices to take advantage of the induced locality. To do so, one may introduce two cosine weights to Eq 10 of CosFormer [51]: one in x direction and the other one in y direction to come up with:

$$ Q_{i,m}K_{j,n}^Tcos(i-j)cos(m-n) $$

which can be expanded to:

$$\scriptstyle{ Q_{i,m}K_{j,n}^T \Big(cos(i)cos(j) + sin(i)sin(j)\Big)\Big(cos(m)cos(n) + sin(m)sin(n)\Big)}$$

which can be regrouped to: 

\begin{align*}
=& \Big(Q_{i,m}cos(i)cos(m)\Big)\Big(K_{j,n}^Tcos(j)cos(n)\Big) 
\\
+& \Big(Q_{i,m}cos(i)sin(m)\Big)\Big(K_{j,n}^Tcos(j)sin(n)\Big) \\
+& \Big(Q_{i,m}sin(i)cos(m)\Big)\Big(K_{j,n}^Tsin(j)cos(n)\Big)  \\
+& \Big(Q_{i,m}sin(i)sin(m)\Big)\Big(K_{j,n}^Tsin(j)sin(n)\Big)
\nonumber
\end{align*}

Hence, for every attention value, CosFormer needs 4 dot products between Q and K vectors while our method needs only one dot product. Hence, following Eq. 12 of the CosFormer paper, CosFormer needs 4 times more FLOPS compared to our method in calculating the attention values (multiplying $Q$, $K$, and $V$ matrices).

\end{document}